\documentclass{article}
\pdfoutput=1

\usepackage{arxiv}

\usepackage[utf8]{inputenc} % allow utf-8 input
\usepackage[T1]{fontenc}    % use 8-bit T1 fonts
\usepackage{hyperref}       % hyperlinks
\usepackage{url}            % simple URL typesetting
\usepackage{booktabs}       % professional-quality tables
\usepackage{amsfonts}       % blackboard math symbols
\usepackage{nicefrac}       % compact symbols for 1/2, etc.
\usepackage{microtype}      % microtypography
\usepackage{lipsum}
\usepackage{amsmath}
\usepackage{graphicx}
\newcommand{\etal}{\emph{et al.}}
\usepackage{caption}
\usepackage{soul}
\graphicspath{ {./images/} }

\title{Anomaly Detection in Satellite Videos using Diffusion Models}

\author{
 Akash Awasthi \\
  Department of Electrical and Computer Engineering\\
  Houston, TX 77004\\
  \texttt{akashcseklu123@gmail.com} \\
   \And
 Son Ly \\
  Department of Electrical and Computer Engineering\\
  University Of Houston\\
  Houston, TX 77004 \\
  \texttt{stly@uh.edu} \\
  \And
  Jaer Nizam \\
  Natural Sciences and Mathematics\\
  University Of Houston\\
  Houston, TX 77004 \\
  \texttt{jaernizam2@gmail.com} \\
  \And
 Samira Zare \\
  Department of Electrical and Computer Engineering\\
  University of Houston\\
  Houston, TX 77005 \\
  \texttt{szare836@uh.edu} \\
    \And
 Videet Mehta \\
  Math and Science Academy\\
  Dulles High School\\
  Sugar Land, TX 77479 \\
  \texttt{mvideet@gmail.com} \\
    \And
 Safwan Ahmed \\
  Natural Sciences and Mathematics\\
  University of Houston\\
  Houston, TX 77004 \\
  \texttt{safwanahmad2002@gmail.com} \\
    \And
 Keshav Shah \\
  School of Engineering\\
 Rice University\\
  Houston, TX 77005 \\
  \texttt{keshav@rice.edu} \\
    \And
 Ramakrishna Nemani \\
  Bay Area Enviornmental Research Institute\\
  NASA AMES \\
  California, CA 94952 \\
  \texttt{nemani@baeri.org } \\
      \And
 Saurabh Prasad \\
  Department of Electrical and Computer Engineering\\
  University of Houston\\
  Houston, TX 77004 \\
  \texttt{sprasad2@uh.edu} \\
      \And
 Hien Van Nguyen \\
   Department of Electrical and Computer Engineering\\
  University of Houston\\
  Houston, TX 77004 \\
  \texttt{hvnguy35@central.uh.edu} \\
}

\begin{document}
\maketitle
\begin{abstract}
  The definition of anomaly detection is the identification of an unexpected event. Real-time detection of extreme events such as wildfires, cyclones, or floods using satellite data has become crucial for disaster management. Although several earth-observing satellites provide information about disasters, satellites in the geostationary orbit provide data at intervals as frequent as every minute, effectively creating a video from space.  There are many techniques that have been proposed to identify  anomalies in surveillance videos; however, the available datasets do not have dynamic behavior, so we discuss an anomaly framework that can work on very high-frequency datasets to find very fast-moving anomalies. In this work, we present a diffusion model which does not need any motion component to capture the fast-moving anomalies and outperforms the other baseline methods.
\end{abstract}

% keywords can be removed
%\keywords{First keyword \and Second keyword \and More}

\section{Introduction}
Greenhouse gas emissions have caused an increase in extreme weather events and climate disasters, resulting in billion-dollar losses \cite{noaabillions}. Wildfires have become a serious issue as well with greenhouse emissions contributing to their severity. These fires destroy vast areas of forests and natural habitats, leading to the loss of biodiversity and wildlife. Additionally, wildfires contribute to soil erosion and degradation, and their smoke and ash can cause air and water pollution has well. Satellites in orbit have proven to be incredibly valuable for disaster management. They can provide valuable images and videos that can help predict and manage wildfires. However, a major issue with these satellites is the significant latency in relaying their data. The data obtained from these satellites are sometimes 2-3 hours past the location of a disaster event \cite{noaabillions}. Consequently, it would limit their effectiveness in the early detection of fast-developing events.

Wildfires are usually unpredictable phenomena; however, there have been many efforts to utilize this satellite data to combat wildfires. Early models use handcrafted thresholds to detect fire pixels \cite{flooding1981identification,kaufman1990remote,pereira1993spectral}, but this is only sufficient for specific regional and seasonal conditions. With the rapid growth of deep learning algorithms, recent models have attempted to effectively utilize satellite data through Convolutional Neural Networks (CNNs). For instance, Phan \etal \cite{phan2019remote} used 3D-CNNs to learn the spatial and spectral patterns of streaming images. Vani \etal \cite{vani2019deep} employed a transfer learning technique to perform fire versus non-fire classification. The Fully Convolutional Network (FCN) \cite{larsen2021deep} is also proposed to segment smoke pixels from non-smoke pixels. Although these models yield exceptional performance in fire detection, these tools are only effective once a fire has grown to a sufficient size that is detectable by the satellites. As a result, these models have a major limitation since the ultimate objective of fire detection is to hinder the propagation of wildfires.

In this paper, we aim to present a novel approach to address the limitations of CNN-based methods. We leverage a class of state-of-the-art generative models, namely diffusion models \cite{sohl2015deep,ho2020denoising,beatgan,voleti2022mcvd,ho2022video}. Diffusion models are developed based on the properties of partial differential equations and Brownian motion to generate samples by iteratively transforming a simple noise distribution into the target distribution. Specifically, we utilize diffusion models to learn the prior distribution and generate high-quality data of normal events. We can then identify the initiation of anomalous events such as fire or smoldering if the satellite data deviates from the prior learned distribution. The results indicate that our proposed method is able to detect small wildfires, which may rapidly grow into major fires within minutes, with high accuracy and a low false positive rate.

Our approach can contribute to the development of effective and timely wildfire detection methods in order to mitigate their devastating impacts. Compared to the previous CNN-based methods, this approach does not require data from high-temporal-resolution satellite videos. It is important to note that there are other works that utilize Generative Adversarial Networks (GANs) \cite{ganfire} to synthesize missing or corrupted data for wildfire detection. However, this paper employs the diffusion model to generate samples for the purpose of detecting wildfires and satellite anomalies (generating-to-detecting), which has not been previously explored in the literature. Furthermore, compared to GANs, diffusion models can generate more diverse and realistic samples \cite{dhariwal2021diffusion}, which can improve the accuracy and robustness of our wildfire detection algorithm. 
%not 100% about the last citation of the paragraph above
Moreover, we note that previous studies rely on satellite datasets such as Landsat-8 \cite{landsat8}, Himawari-8 \cite{na2018himawari}, MODIS Collection 6 (MOD/MYD14) \cite{GiglioMODIS}, and VIIRS 375m (VNP14IMG) \cite{chen2022california}, which are designed for fire detection and segmentation purposes. However, these existing datasets fail to meet our model's specifications, as the images within these datasets are predominantly conventional fire events rather than anomalous occurrences. If we were to train our model using these datasets, it would primarily learn to recognize standard fire patterns like the presence of flames, smoke, or thermal signatures. Regrettably, it would lack the ability to effectively predict atypical fire events such as smoldering, which is the objective we are attempting to tackle. Therefore, we have constructed our own dataset specifically for fire anomaly detection based on the data from the GOES-16 and 17 satellites operated by the NOAA. These satellites use technology that allows them to collect reflected and emitted radiation from the Earth's surface. This allows us to capture less visible events such as smoldering and other anomalies, which are the target of our dataset.  

In this paper, we first acknowledge the related works in section \ref{related_works}. Then, we provide a detailed description of our approach and diffusion models in section \ref{method}. Section \ref{experiment} presents experimental results on our real-world satellite data to demonstrate the effectiveness of our method in detecting small wildfires with a high accuracy and a low false positive rate.

\section{Related Works}
\label{related_works}

\subsection{Fire Detection/Tracking on Satellite Data}
Satellite images and videos have been extensively used to fight wildfires, with various studies focusing on active fire detection and tracking. To address the limitations of the early thresholding models mentioned in the introduction \cite{flooding1981identification,kaufman1990remote,pereira1993spectral}, dynamic thresholding techniques have been utilized to adapt to local contextual conditions and minimize false alarms for smaller and cooler fires \cite{flasse1996contextual,zhang2017approaches,xu2017major,di2018geostationary}. Contextual algorithms remain the most common approach for active fire detection due to their computational efficiency \cite{wooster2021satellite}. Incorporating time constraints upon the dynamic thresholds can also reduce false alarms \cite{roberts2014development,filizzola2016rst}.

Recent advancements in deep learning have enabled greater levels of exploration beyond manually designed operators. For example, a state-of-the-art study \cite{rostami2022active} proposed a CNN-based network consisting of different convolution kernel sizes to detect fires of varying sizes and shapes. Other researchers have tried using handcrafted features to improve fire tracking \cite{xu2017real,na2018himawari} or developing algorithms that analyze the brightness of infrared images and the offset of the sunrise to the thermal sunrise time of a non-fire condition \cite{udahemuka2020characterization}. 

While fire detection and tracking are important, this paper focuses on the fire anomaly detection approach, which can detect fire events even when the fire is not visible in the images, such as smoldering. This approach uses diffusion models to learn the prior distribution and generate useful data on non-fire events. Thus, when asked to generate an image of an anomaly event, such as fire or smoldering, the diffusion model is capable of producing good result. At the time of publishing, this is the first paper (to the best of our knowledge) that fights wildfire by taking advantage of the anomaly detection task using video diffusion models.

\subsection{Video Diffusion Models}
Diffusion models are high-performing, likelihood-based models commonly used in synthetic image generation. Diffusion models designed for image generation consist of a U-Net architecture and perform better than GANs, specifically in image resolutions greater than $64 \times 64$. The GAN's ability to capture diversity is relatively poor compared to the diffusion model \cite {dhariwal2021diffusion}. However, these benefits come at the expense of longer computation times due to denoising steps and lower fidelity metrics. Additionally, model collapse and non-convergence further reduce the image generation qualities in a GAN \cite{haiyangchallenge}.  Studies have focused primarily on using diffusion models for image and audio generation; however, developing synthetic video generation algorithms is a rising interest that has sprouted from evaluating diffusion models on different data modalities \cite{ho2022video}. 
 
Ho et al. \cite{ho2022video} proposed the creation of a video diffusion model which used the reconstruction-guidance sampling method to approximate the conditional distributions. The Residual Video Diffusion (RVD) model, was proposed by Yang et al. \cite{yang2022diffusion} for video prediction. The RVD model uses a residual-based approach to model the difference between predicted and true video frames and is effective at modeling the conditional distribution of future video frames given past frames as input. Video Implicit Diffusion Models (VIDM) employ separate content and motion generation streams for artificial frame generation\cite{mei2022vidm}. The content stream utilizes a modified U-Net architecture to model the distribution of video frames, while the motion stream models the changes in motion over a sequence of random frames. Further research is necessary to address the issue of discontinuous motion in VIDM. 
VideoGPT \cite{yan2021videogpt} is a novel transformer-based model used for compression and reconstruction, distinct from traditional diffusion models. This model leverages a fusion of a vector quantized variations autoencoder (VQ-VAE) and GPT architecture to efficiently compress data into a dense, discretized latent space, which is then utilized for image reconstruction. This approach  enhances the computational efficiency of the system while maintaining similar evaluation metrics to the GAN.

%##################################################
\section{Methodology}
\label{method}

\subsection{Diffusion Models}
Diffusion models are a class of generative models that aim to model the probability distribution of a dataset. They operate by employing an iterative process called diffusion, which allows them to capture the underlying data distribution. By learning the conditional probabilities of the data based on its previous states and the applied noise at each diffusion step, diffusion models gain the ability to generate new samples that resemble the training data. The denoising diffusion probabilistic models, which are the type of diffusion models used in this paper, can be decomposed into two distinct processes: the forward process, where the training data is progressively corrupted by Gaussian noise, and the reverse process, which generates new samples from the original distribution by following the reverse steps.

The forward process can be modeled through the following equation \cite{croitoru2023diffusion}:

\begin{align*}
    p(x_t|x_{t-1}) = \mathcal{N}\left(x_t; \frac{x_{t-1}}{\sqrt{1-\beta_t}}, \beta_t I\right), \quad \forall t \in \{1, \ldots, T\}
\end{align*}

This formula represents the conditional probability distribution of $x_t$ given $x_{t-1}$ in the forward process. It specifies that $x_t$ is sampled from a normal distribution with mean $\frac{x_{t-1}}{\sqrt{1-\beta_t}}$ and covariance $\beta_t I$, where $\{\beta_1, \ldots, \beta_t\}$ are the hyper-parameters representing the variance schedule across diffusion steps, and $I$ is the identity matrix with the same dimensions as the input image $x_0$.

In the reverse process, the model tries to minimize a variational lower bound of the negative log-likelihood. The objective, denoted as \(L_{\text{vlb}}\), is given by the following formulation\cite{croitoru2023diffusion}:
\begin{align}
\begin{split}
L_{\text{vlb}} = &-\log p_\theta(x_0|x_1) + \text{KL}(p(x_T|x_0) \| \pi(x_T)) \\
&+ \sum_{t>1} \text{KL}(p(x_{t-1}|x_t, x_0) \| p_\theta(x_{t-1}|x_t))
\end{split}
\end{align}

In this formulation, \(\text{KL}\) denotes the Kullback-Leibler divergence between two probability distributions. The first term represents the negative log-likelihood of predicting \(x_0\) given \(x_1\). The second term measures the KL divergence between the distribution of \(x_T\) given \(x_0\) and a standard Gaussian distribution \(\pi(x_T)\). The third term sums over the KL divergences between the true posterior distribution \(p(x_{t-1}|x_t, x_0)\) and the predicted distribution \(p_\theta(x_{t-1}|x_t)\) at each time step \(t\) of the reverse process.

Using this equation, the diffusion model attempts to approximate the reverse process and generate new samples from the original distribution \(p(x_0)\) by starting from a sample \(x_T \sim N(0, I)\) and following the reverse steps.

Diffusion models are particularly effective in capturing complex dependencies and generating realistic samples. They can be applied to various domains, including images, videos, text, and other structured or unstructured data types. The generated samples can be used for tasks such as data synthesis, data augmentation, or generating new data points for downstream applications\cite{croitoru2023diffusion}.

% Similar to Section 2 in \href{https://arxiv.org/pdf/2205.09853.pdf}{MCVD: Masked Conditional Video Diffusion}
\subsection{Conditioned Generation}
Conditioned generation is a type of generative modeling in which a model is trained to generate output samples that satisfy certain conditions or constraints. The goal is to train a model that can generate high-quality output samples that not only looks realistic but also satisfies the specified conditions or constraints. In the scope of computer vision and image synthesis, we will explore two types of conditioned generation models: Classifier Guided Diffusion and Classifier-Free Guidance.
\vspace{-.3cm}
\subparagraph{Guided Diffusion}To improve the quality of generated images, a classifier is utilized to provide information to the diffusion model about the desired target distribution. This classifier, as described in \cite{dhariwal2021diffusion}, takes the form of $f_{\phi}(y|\mathbf{x}_t,t)$ where $\mathbf{x}_t$ represents the noisy image. By using gradients in the form of $\nabla{\mathbf{x}} \log f{\phi}(y|\mathbf{x}_t)$, the diffusion sampling process is guided towards the target image by modifying the noise prediction of the original model. To achieve this, we recall the following equation \cite{song2019generative}:
\begin{align}
\nabla_{\mathbf{x}_t} \log q(\mathbf{x}_t) = - \frac{1}{\sqrt{1 - \bar{\alpha}t}} 
 \boldsymbol{\epsilon}\theta(\mathbf{x}_t, t)
\end{align}

\noindent Using this, we can write the score function for the joint distribution $q(\mathbf{x}_t, y)
$ as follows:
\begin{align}
\nabla_{\mathbf{x}_t} \log q(\mathbf{x}_t, y) &= \nabla{\mathbf{x}_t} \log q(\mathbf{x}_t) + \nabla{\mathbf{x}_t} \log q(y \vert \mathbf{x}_t)
\end{align}

\noindent From this, we can derive the new classifier-guided predictor in the form:
\begin{align}
\bar{\boldsymbol{\epsilon}}_\theta(\mathbf{x}_t, t) = \boldsymbol{\epsilon}_\theta(x_t, t) - \sqrt{1 - \bar{\alpha}_t} \nabla_{\mathbf{x}_t} \log f_\phi(y \vert \mathbf{x}_t)
\end{align}
where $\bar{\boldsymbol{\epsilon}}_\theta$ is the modified noise prediction.

\subparagraph{Free Guidance Diffusion} Contrasting with guided diffusion, free guidance diffusion does not rely on an external classifier. Instead, the contents of the image itself guide the diffusion process, which is aided by a diffusion term in the generative model. The training process involves using both a conditional model $p_\theta(\mathbf{x} \vert y)$ and an unconditional denoising diffusion model $p_\theta(\mathbf{x})$ \cite{ho2022classifier}. An implicit classifier is used for training, where conditioning information is periodically discarded at random to allow the model to generate images unconditionally. The gradient for the implicit classifier can be derived from the conditional and unconditional score estimators using the following equation:

\vspace{-.09cm}
\begin{align}
\nabla_{\mathbf{x}_t} \log p(y \vert \mathbf{x}t)
&= \nabla{\mathbf{x}_t} \log p(\mathbf{x}t \vert y) - \nabla{\mathbf{x}_t} \log p(\mathbf{x}_t)
\end{align}

\subsection{Conditional diffusion for video}
Given $\boldsymbol{x}_0 \sim q(x)$, the forward process corrupts $\boldsymbol{x}_0$ with small amount of Gaussian noise at each time stamp $t \in [0, T]$ that satisfies Markovian transition:
%##################################################
\begin{align}
q(\boldsymbol{x}_t|\boldsymbol{x}_{t-1}) &= \mathcal{N}(\boldsymbol{x}_t, \sqrt{1-\beta_t}\boldsymbol{x}_{t-1},\beta_t\boldsymbol{I})  \\
q(\boldsymbol{x}_{1:T}|\boldsymbol{x}_0) &= \prod_{t=1}^{T}q(\boldsymbol{x}_t|\boldsymbol{x}_{t-1}) 
\end{align}
%##################################################
The $x_0$ is gradually degraded as the step becomes larger and eventually $x_T$ is equivalent to an isotropic Gaussian distribution. A `nice property' of this process is that, from $x_0$, $x_t$ can be sampled at any arbitrary $t$ using the re-parameterization trick \cite{vae} as:
%##################################################
\begin{align}
    q_t(\boldsymbol{x}_t|\boldsymbol{x}_0) &= \mathcal{N}(\boldsymbol{x}_t;\sqrt{\bar{\alpha}}\boldsymbol{x}_0,(1-\bar{\alpha})\boldsymbol{I})  \\
\boldsymbol{x}_t &= \sqrt{\bar{\alpha}}\boldsymbol{x}_0 + \sqrt{1-\bar{\alpha}_t}\boldsymbol\epsilon 
\end{align}
%##################################################
where $\bar{\alpha}_t = \prod_{i=1}^{t}(1 - \beta_i)$, and $\boldsymbol\epsilon \sim \mathcal{N}(\boldsymbol 0, \boldsymbol I)$.\\
To reverse the above process and generate new samples, we need to approximate $q(\boldsymbol{x}_{t-1}|\boldsymbol{x}_t)$ by learning an $p_\theta$, which are tractable when conditioned on $\boldsymbol{x}_0$:
%##################################################
\begin{align}
q(\boldsymbol{x}_{t-1}|\boldsymbol{x}_t,\boldsymbol{x}_0)&=\mathcal{N}(\boldsymbol{x}_{t-1}; \tilde{\boldsymbol{\mu}}_t (\boldsymbol{x}_t,\boldsymbol{x}_0),\tilde{\beta}_t\boldsymbol I)   \\
\text{where}\quad \tilde{\boldsymbol{\mu}_t}(\boldsymbol x_t, \boldsymbol x_0) &= \frac{\sqrt{\bar\alpha_{t-1}}\beta_t }{1-\bar\alpha_t}\boldsymbol x_0 + \frac{\sqrt{\alpha_t}(1-\bar\alpha_{t-1})}{1-\bar\alpha_t} \boldsymbol x_t \quad   \\
\text{and}\quad \tilde\beta_t &= \frac{1-\bar\alpha_{t-1}}{1-\bar\alpha_t}\beta_t 
\end{align}
%##################################################
Thanks to the `nice property', we can estimate $\hat{\boldsymbol{x}}_0 = (\boldsymbol{x}_t - \sqrt{1-\bar\alpha_t} \boldsymbol\epsilon_t)/\sqrt{\bar\alpha_t}$. Since $\boldsymbol x_t$ is available from the forward process, we can re-parameterize the Gaussian noise term instead to predict $\boldsymbol\epsilon_t$ by $p_\theta$. Thus, the loss term would be:
%##################################################
\begin{align}
    L(\theta) = \mathbb{E}_{t,\boldsymbol x_0,\boldsymbol\epsilon}\left [ \left \|\boldsymbol\epsilon - \boldsymbol\epsilon_\theta (\sqrt{\bar{\alpha_t}}\boldsymbol{x}_0 + \sqrt{1-\bar\alpha_t} \boldsymbol\epsilon \:|\: t)  \right \|_2^2 \right ]
\end{align}
%##################################################
\subsection{Anomaly Prediction Via Conditional Diffusion}
Anomaly detection is the identification of abnormal events which are not expected. Video prediction can be used to identify anomalies in the data by predicting the future frame and comparing it with the ground truth. Our diffusion-based method uses the video prediction task to identify the anomalies. The diffusion based method does not need any external motion component to capture the high-quality motion and it works for high-frequency and dynamic datasets (satellite data) to identify the anomalies. We can identify the anomalies which are caused due to motion as well as color. 
\subsection{Video Prediction using Conditional Diffusion}
\begin{figure}[htp]
    \centering
    \includegraphics[width=\linewidth]{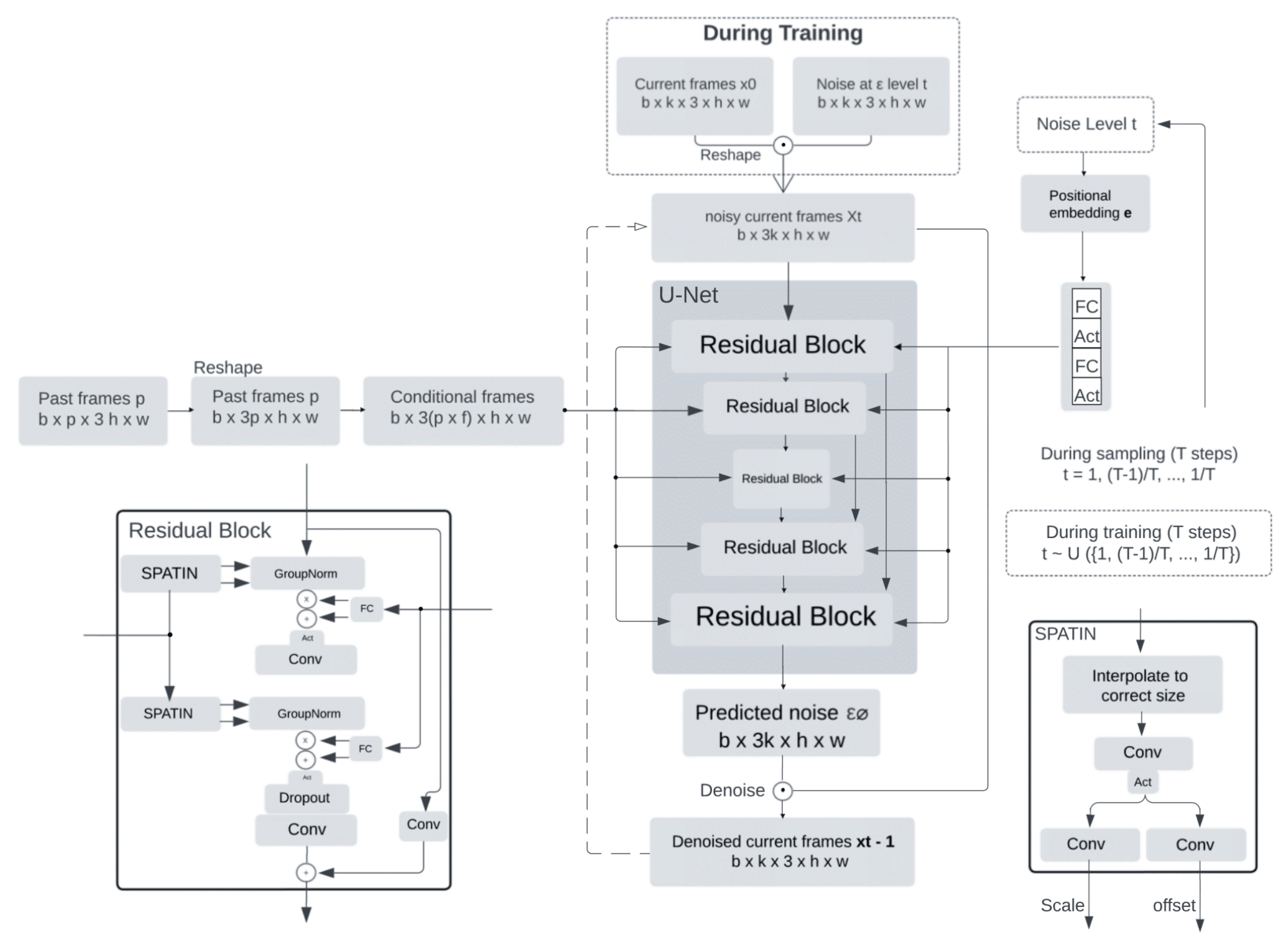}
    \captionsetup{justification=raggedright, singlelinecheck=false}
    \caption{A U-Net model is provided with noisy current frames where the residual blocks incorporate information from fast and future frames. The U-Net model predicts the noise present in the current frames which is used to denoise the current frame.}
    \label{fig:arch}
\end{figure}

We model the conditional distribution of the video frames by incorporating the past frames. But we also perform an ablation experiment by conditioning the past and future frames. Suppose there are $p$ {} past frames and $k$ current frames. We condition the diffusion model on the past frames to predict future frames. 
\begin{equation}\label{eq:vidprecond}
\small
    L(\theta)_{vidpred} = \mathbb{E}_{t,[\boldsymbol p ,\boldsymbol x_0],\boldsymbol\epsilon}\left [ \left \|\boldsymbol\epsilon - \boldsymbol\epsilon_\theta (\sqrt{\bar{\alpha_t}}\boldsymbol{x}_0 + \sqrt{1-\bar\alpha_t} \boldsymbol\epsilon \:|\: \boldsymbol p,t)  \right \|_2^2 \right ]
\end{equation} \\
After training the diffusion model, in the first window of $p+k$ frames, we predict the $k$ frames conditioned on the past $p$ frames and then shift the window to the $p+k$ frames and then repeat the same process for the whole video. For modeling $\epsilon_\theta$ shown in Eq. (\ref{eq:vidprecond}) we use variants of networks shown in Fig. \ref{fig:arch}.

\begin{figure}[t]
    \centering
    \includegraphics[width=\linewidth]{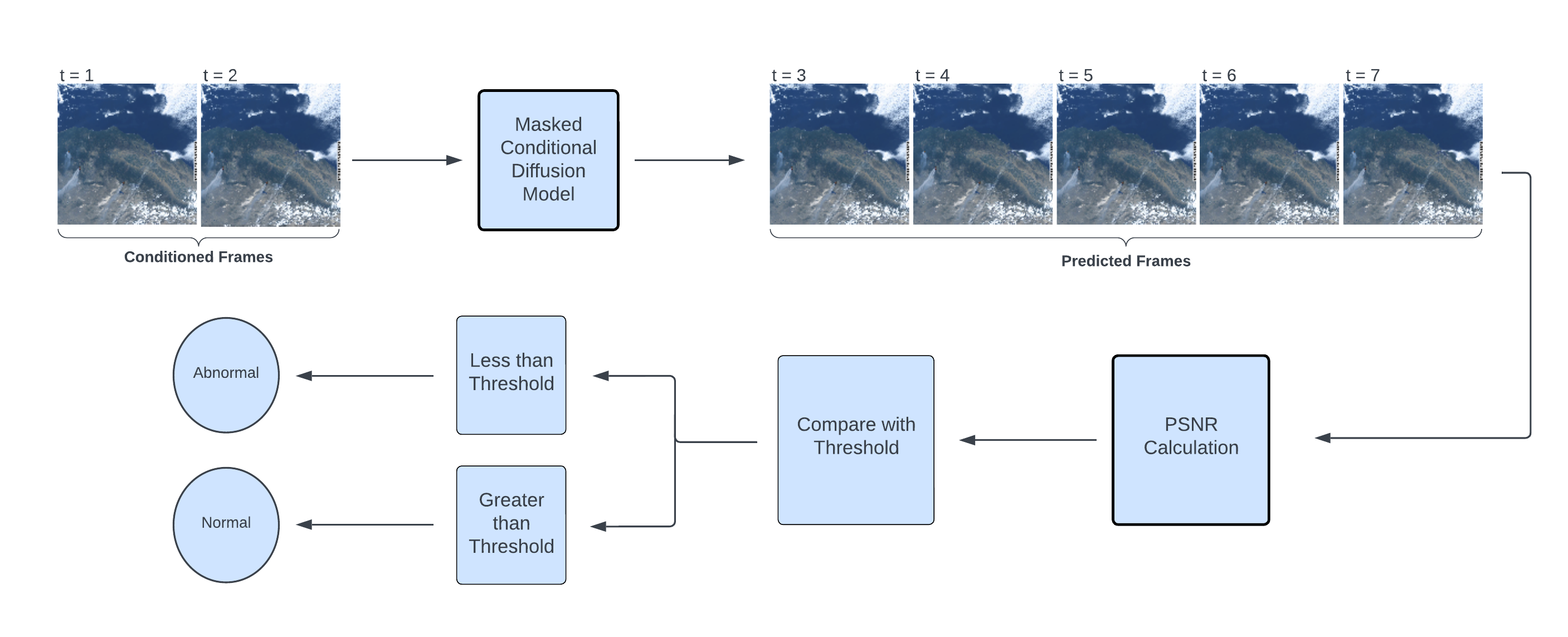}
    \caption{Strategy to predict the future frame with the sliding window approach}
    \label{fig:strategy}
\end{figure}

\subsection{Architecture}
\subparagraph{U-Net}
Due to its effectiveness and efficiency, U-Net is a popular convolutional neural network architecture for conducting image segmentation tasks. U-Net partitions an input image into multiple classes of pixels in a process known as semantic segmentation. As explained by Ronneberger et al.\cite{ronneberger2015u}, the network consists of two parts: a contracting path and an expansive path. The contracting path applies repeated convolutions and max pooling operations to downsample the image and capture contextual information, while the expansive path uses upsampling and skip connections to reconstruct the original image resolution while preserving the contextual information. The U-Net model uses skip connections to combine fine-grained details and structures in an image (local information) with the overall context and spatial relationships in an image (global information). This is what allows a U-Net model to effectively segment images with fine details and irregular shapes.

\subparagraph{Residual Block}
Primarily used in deep learning networks such as convolutional neural networks, residual blocks provide a solution to the degradation problem, where the accuracy of the network starts to degrade as the depth of the network increases. Residual blocks implement a way for the network to model the differences between the input and output of each block making it easier for the network to optimize deeper layers. Residual Blocks (RBs) use a method dealing with shortcut connections that add the input of a block directly to its output to allow information to flow through the network more easily and improve the gradient flow during back-propagation. Residual blocks have been widely adopted in many state-of-the-art deep learning architectures and have achieved outstanding performance on various computer vision tasks, such as image recognition, object detection, and semantic segmentation.\cite{he2016deep}

We have used the architecture from the model proposed for the video prediction \cite{voleti2022mcvd}. U-Net backbone is used as the denoising network with some changes. This architecture uses multi-head self-attention and 2D convolution and adaptive group normalization \cite{wu2018group}. Position encoding is used for the noise level and it is processed using transformer-style encodings.

 We use past frames as conditioned frames and concatenate along the channel dimension. Current noisy frames are created in the forward diffusion process and used the timestep t frames with noise as input in the denoising network. Concatenated conditional frames are passed through the network  that affects  the conditional normalization also known as   Spatially-Adaptive (DE)normalization (SPADE) \cite{cadena2021spade}. This SPADE module accounts for time and motion. The satellite  dataset which we use is very dynamic in nature since the clouds move very fast in 5 minutes. Therefore, this inbuilt module helps to capture fast motion. This is better than using FlowNet which increases the computational complexity.
 \begin{equation}
    \textbf{e}(t)=\left[ \ldots,\cos \left(t c^{\frac{-2d}{D}} \right),  \sin \left(t c^{\frac{-2d}{D}} \right) , \ldots  \right]^{\mathrm{T}},
\end{equation}\\
where $d = 1,...,D/2$ , $D$ is the number of dimensions of the embedding, and $c = 10000$. Each embedding vector is passed through a fully connected layer with an activation function and then another fully connected layer

\section{Experiments}
\label{experiment}
We have used the diffusion model architecture  for the video prediction task. We use the 2 past frames as conditioned frames and predict 5 frames at a time and then shift to the next window. For sampling, we use the DDPM sampling \cite{ho2020denoising} with the 100 sampling steps with the model being trained with these 100 sampling steps.

\subsection{Datasets}
We used data from geostationary satellites that are synchronized with Earth’s spin to hover over the same point on Earth making them ideal for monitoring environmental dynamics. The GOES-16 and 17 satellites covering the US carry the Advanced Baseline Imager that collects reflected and emitted radiation from the Earth in 16 wavelength bands. Three data products are available from NOAA: the entire Northern Hemisphere every 15 minutes, the Continental US at 5 minutes, and the mesoscale user-directed (1000km x 1000km) at every minute. In this proof of concept study, we used the 5-minute data. For the current experiment, we have used the data from the Northern California region and divided the videos into short clips having 14 frames each. 

Our model is trained on the videos containing the normal frames  and tested on the mixture of  normal and abnormal frames containing the fires. The main challenge  here is to extract the video clips containing the normal frames. It can be done manually by identifying the clips from the pool of the video dataset, but we have used the pre-trained YOLOv5 \cite{yolov5} to extract the abnormal and normal frames

YOLOv5 is trained on the publicly available fire images and we use this pre-trained model to identify the short clips having fire. This helps us to identify the clips  which do not contain any fire, smoke, or fog and these can be used as normal datasets for training the anomaly detection framework. We have extracted around 500 normal videos and 20 abnormal videos containing fire, smoke, and fog. Anomalous images are shown below in Fig. \ref{fig:abnormalframe}.
 
\begin{figure}[htp]
    \centering
    \includegraphics[width=\linewidth]{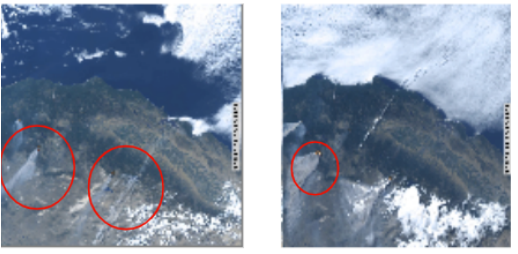}
    \caption{Abnormal Frames}
    \label{fig:abnormalframe}
\end{figure}

Real frames of the videos are very high resolution (4k) but we have resized each frame of the video to 128x128.

\subsection {Anomaly Detection on Testing Data}
We have trained our model with the normal videos having no anomalous event so we assume that the model can predict the normal images well. MSE is used to calculate the difference between the predicted image and the true image but \cite{mathieu2015deep} shows that PSNR (Peak Signal to Noise Ratio)  is an efficient way to access the image quality. 
\begin{equation}
    PSNR(I,\widehat{I})=10\:log_{10}\frac{[max_{\widehat{I}}]^2}{\frac{1}{N}\sum_{N}^{i=0}(I_i-\widehat{I}_i)^2}
\end{equation}
For anomalous frames, the PSNR value is high, but for the normal frames the PSNR is low. We also normalize the PSNR of each testing video between 0 and 1, following this work \cite{medel2016anomaly}. This normalized PSNR is called the regular score.

\begin{equation}
    S(t)=\frac{PSNR(I_t,\widehat{I}_t)-min_t\:PSNR(I_t,\widehat{I}_t)}{max_t\:PSNR(I_t,\widehat{I}_t)-min_t\:PSNR(I_t,\widehat{I}_t)}
\end{equation}
Based on this regular score we can predict whether the frame is normal or abnormal. We can set the threshold to distinguish between normal or abnormal frames.

\section{Results and Discussions}

\subsection {Evaluation Matrices}
In previous literature on anomaly detection, the ROC curves have been established as the primary metric for algorithm performance. This is typically done by gradually adjusting the threshold for regular scores to identify anomalies\cite{mahadevan2010anomaly}. To assess the performance of the anomaly detection algorithm, the area under the ROC curve (AUC) is calculated. A higher AUC indicates better performance in distinguishing between anomalous and regular events.
% Please add the following required packages to your document preamble:
% \usepackage{booktabs}
\begin{table}[t]
\centering
\begin{tabular}{@{}llc@{}}
\toprule
& Method                  & AUC (\%)    \\ \midrule
& Future frame Prediction\cite{liu2018future} & 73.2 \\
& ConvLSTM-AE \cite{burgeLSTM}             & 71.5 \\
& MLEP                    & 78   \\
& Conv-AE                 & 68   \\
& Diffusion               & 80.3 \\ \bottomrule
\end{tabular}
\caption{AUC Score Comparison}
\label{tab:aucresult}
\end{table}

As is shown in  Table \ref{tab:aucresult}, we compare our method with the other deep learning baselines and our method outperforms the other methods in terms of AUC score. 
\subsection {PSNR Plot for the Normal and Abnormal Video}
\begin{figure}[htp]
    \centering
    \includegraphics[width=\linewidth]{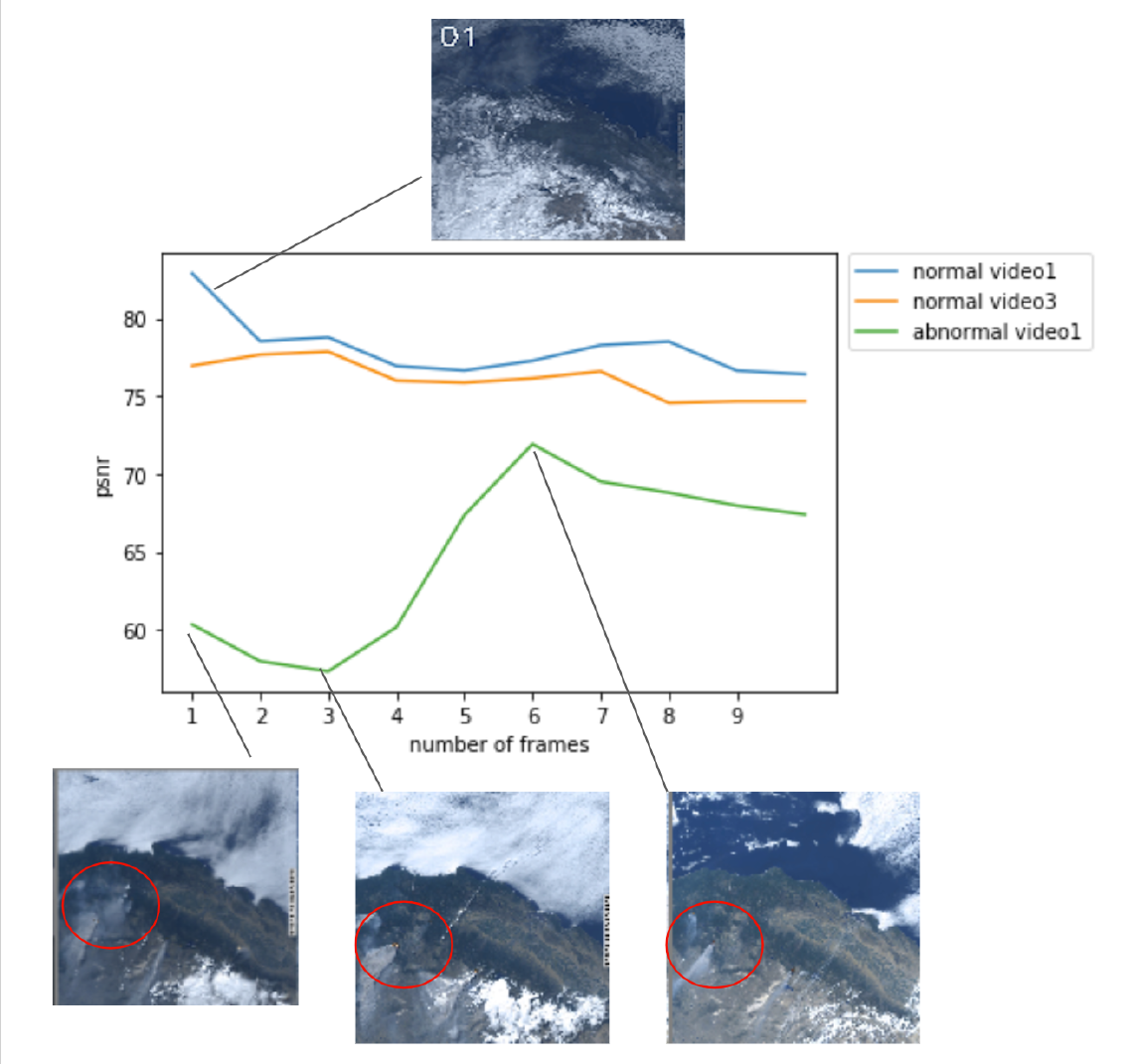}
    \caption{PSNR curves for Normal and Abnormal Videos}
    \label{fig:psnrcurves1}
\end{figure}

\begin{figure}[htp]
    \centering
    \includegraphics[width=\linewidth]{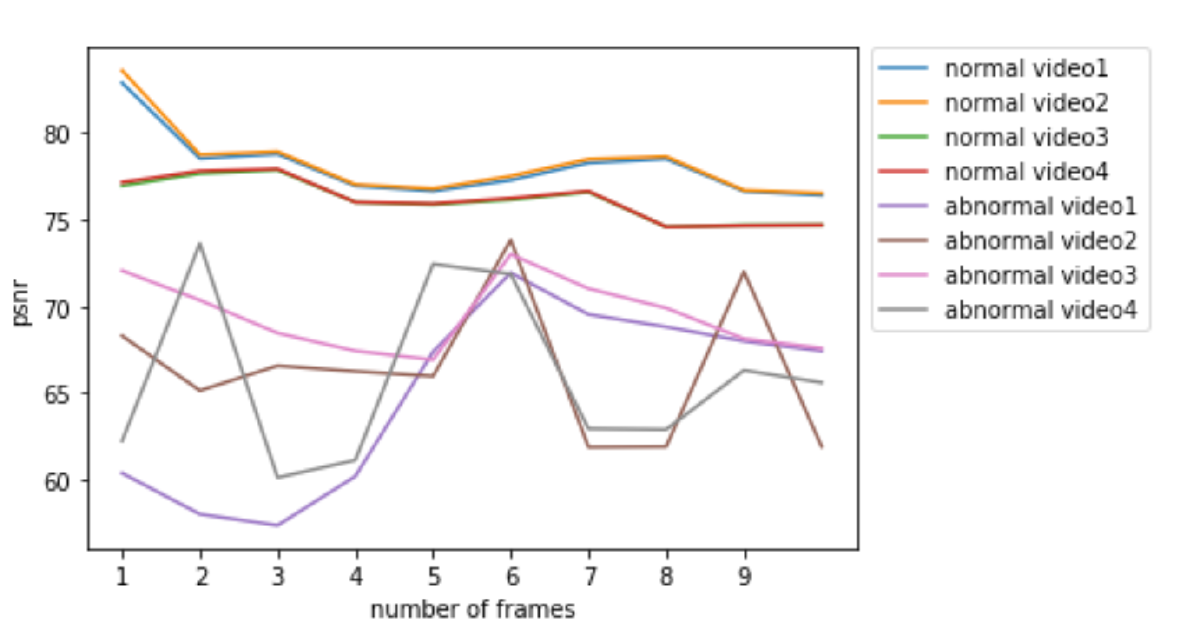}
    \caption{PSNR plot for normal and abnormal videos with fire, smoke, and fog as anomaly}
    \label{fig:psnrcurves2}
\end{figure}

We have calculated the PSNR as described above for both the normal videos which do not have any fire, smoke, or fog. As shown in Fig. \ref{fig:psnrcurves1} and \ref{fig:psnrcurves2}, the PSNR for those videos is high and continuously high since no anomalous event is identified in the whole video. However, in the case of the \textit{abnormal video1} curve, the initial value of the PSNR is low due to the fire and smoke being very high, and it continues to decrease. In Frame 6, the fire dies, but there is a little bit of smoke resulting in the PSNR of the frame increasing afterward.
\begin{figure}[htp]
    \centering
    \includegraphics[width=\linewidth]{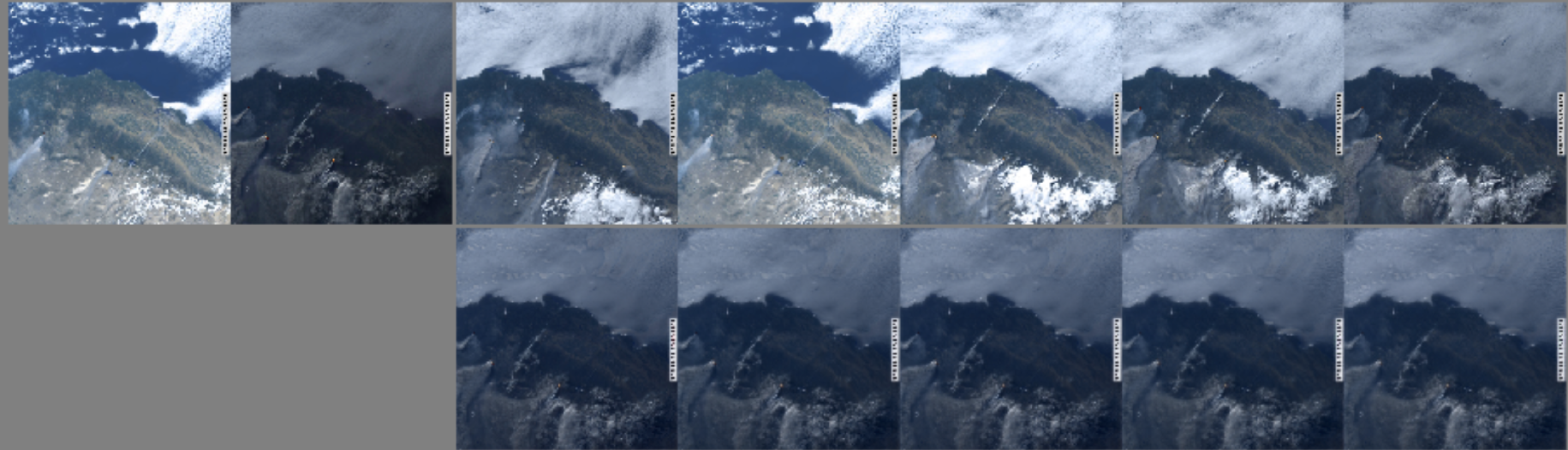}
    \caption{ (Abnormal Video) First row represents the real frames and 2 rows represent the predicted frames  }
    \label{fig:abnormalvid}
\end{figure}

\begin{figure}[htp]
    \centering
    \includegraphics[width=\linewidth]{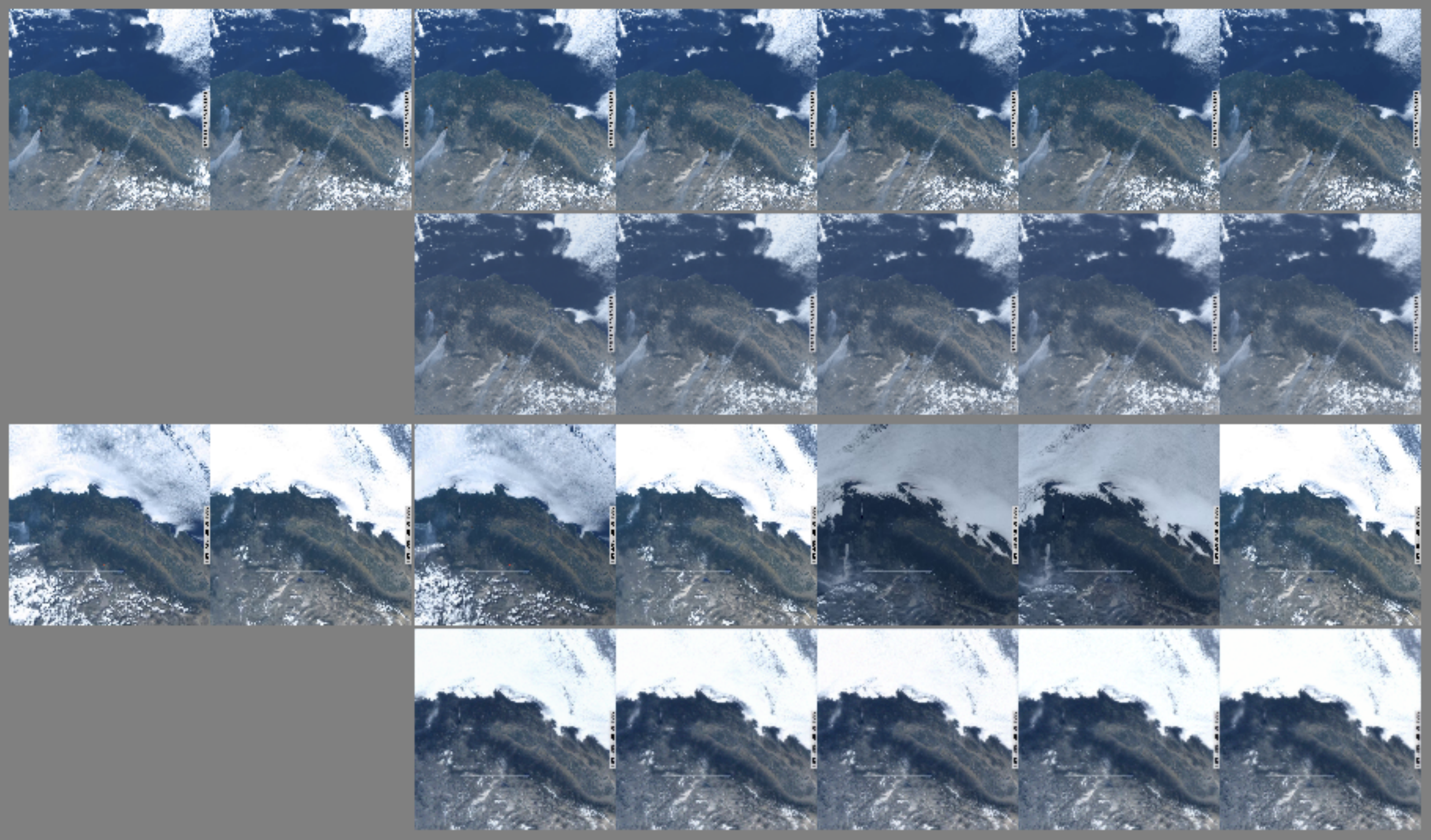}
    \caption{( Normal Video ) First row represents the real frames and 2 rows represent the predicted frames  }
    \label{fig:normalvid}
\end{figure}

Figure \ref{fig:abnormalvid} displays two rows of frames, where the first row represents actual abnormal frames and the second-row exhibits predicted abnormal frames. The first two frames in the first row correspond to conditioned frames stemming from the initial video window, comprising two conditioned frames and five predicted frames. The predicted frames exhibit lower PSNR, as depicted in Figure \ref{fig:psnrcurves1}, and suffer from blurriness, attributable to the absence of fire images in the training data. Therefore, the image quality deteriorates significantly in frames containing fire, smoke, or fog.
 
\subsection{PSNR for Multiple Anomalies Videos}

Above is the PSNR plot for the multiple abnormal and normal videos. The first four  curves for the normal videos report high PSNR and the other four curves for the abnormal videos report low PSNR values since they include abnormalities in their frames. The PSNR varies here since the intensity of the fire and smoke changes with respect to the frame. The corresponding video frames for each abnormal video are attached in the supplementary section.  

We performed further experiments with conditioning on the past and future  frames and the results are provided in Table \ref{tab:abl} and additional predicted images are attached in the supplementary section. Conditioning on both past and future frames did not yield satisfactory results in identifying anomalous frames.

\begin{table}[t]
\centering
\begin{tabular}{@{}llc@{}}
\toprule
& Frames                         & AUC (\%)    \\ \midrule
& 2 Future + 3 Predicted + 2Past & 56.2 \\
& 20 Past +5 predicted           & 70 \\
 \bottomrule
\end{tabular}
\caption{AUC Score Comparison}
\label{tab:abl}
\end{table}

\section{Conclusion}
Wildfire analysis is an effective method for combating extreme climatic events. While fire detection and tracking are undoubtedly important, these tasks are only helpful when the fire is already developed to a certain visualizable degree in captured satellite images. This paper presents a method to detect active fires in terms of anomaly detection tasks using diffusion models. The diffusion model generates a good image for the non-fire event by learning the prior distribution of this type of data; thus, when asked to generate a fire event, the diffusion model generates the highest AUC score compared to all other baseline models. By recognizing these results, it is visible how the proposed method can distinguish between non-fire and fire events and how the empirical results support our findings.

\bibliographystyle{unsrt}  
\bibliography{references}

\begin{thebibliography}{10}

\bibitem{noaabillions}
NOAA National~Centers for Environmental Information~(NCEI).
\newblock U.s. billion-dollar weather and climate disasters (2022).
\newblock {\em DOI: 10.25921/stkw-7w73}, 2022.
\newblock \url{https://www.ncei.noaa.gov/access/billions/}.

\bibitem{flooding1981identification}
River Flooding.
\newblock Identification of subresolution high temperature sources using a
  thermal ir sensor.
\newblock {\em Photogrammetric Engineering and Remote Sensing}, (9), 1981.

\bibitem{kaufman1990remote}
Yoram~J Kaufman, A~Setzer, C~Justice, CJ~Tucker, MC~Pereira, and I~Fung.
\newblock Remote sensing of biomass burning in the tropics.
\newblock In {\em Fire in the tropical biota}, pages 371--399. Springer, 1990.

\bibitem{pereira1993spectral}
MC~Pereira and AW~Setzer.
\newblock Spectral characteristics of fire scars in landsat-5 tm images of
  amazonia.
\newblock {\em Remote Sensing}, 14(11):2061--2078, 1993.

\bibitem{phan2019remote}
Thanh~Cong Phan and Thanh~Tam Nguyen.
\newblock Remote sensing meets deep learning: exploiting
  spatio-temporal-spectral satellite images for early wildfire detection.
\newblock Technical report, 2019.

\bibitem{vani2019deep}
K~Vani et~al.
\newblock Deep learning based forest fire classification and detection in
  satellite images.
\newblock In {\em 2019 11th International Conference on Advanced Computing
  (ICoAC)}, pages 61--65. IEEE, 2019.

\bibitem{larsen2021deep}
Alexandra Larsen, Ivan Hanigan, Brian~J Reich, Yi~Qin, Martin Cope, Geoffrey
  Morgan, and Ana~G Rappold.
\newblock A deep learning approach to identify smoke plumes in satellite
  imagery in near-real time for health risk communication.
\newblock {\em Journal of exposure science \& environmental epidemiology},
  31(1):170--176, 2021.

\bibitem{sohl2015deep}
Jascha Sohl-Dickstein, Eric Weiss, Niru Maheswaranathan, and Surya Ganguli.
\newblock Deep unsupervised learning using nonequilibrium thermodynamics.
\newblock In {\em International Conference on Machine Learning}, pages
  2256--2265. PMLR, 2015.

\bibitem{ho2020denoising}
Jonathan Ho, Ajay Jain, and Pieter Abbeel.
\newblock Denoising diffusion probabilistic models.
\newblock {\em Advances in Neural Information Processing Systems},
  33:6840--6851, 2020.

\bibitem{beatgan}
Prafulla Dhariwal and Alexander Nichol.
\newblock Diffusion models beat gans on image synthesis.
\newblock {\em Advances in Neural Information Processing Systems},
  34:8780--8794, 2021.

\bibitem{voleti2022mcvd}
Vikram Voleti, Alexia Jolicoeur-Martineau, and Christopher Pal.
\newblock Mcvd: Masked conditional video diffusion for prediction, generation,
  and interpolation.
\newblock {\em arXiv preprint arXiv:2205.09853}, 5(4.1):4, 2022.

\bibitem{ho2022video}
Jonathan Ho, Tim Salimans, Alexey Gritsenko, William Chan, Mohammad Norouzi,
  and David~J Fleet.
\newblock Video diffusion models.
\newblock {\em arXiv preprint arXiv:2204.03458}, 2022.

\bibitem{ganfire}
Jose~D Bermudez, Patrick~N Happ, Raul~Q Feitosa, and Dario~AB Oliveira.
\newblock Synthesis of multispectral optical images from sar/optical
  multitemporal data using conditional generative adversarial networks.
\newblock {\em IEEE Geoscience and Remote Sensing Letters}, 16(8):1220--1224,
  2019.

\bibitem{dhariwal2021diffusion}
Prafulla Dhariwal and Alexander Nichol.
\newblock Diffusion models beat gans on image synthesis.
\newblock {\em Advances in Neural Information Processing Systems},
  34:8780--8794, 2021.

\bibitem{landsat8}
Gabriel~Henrique de~Almeida~Pereira, Andre~Minoro Fusioka, Bogdan~Tomoyuki
  Nassu, and Rodrigo Minetto.
\newblock Active fire detection in landsat-8 imagery: A large-scale dataset and
  a deep-learning study.
\newblock {\em ISPRS Journal of Photogrammetry and Remote Sensing},
  178:171--186, 2021.

\bibitem{na2018himawari}
Li~Na, Jiquan Zhang, Yulong Bao, Yongbin Bao, Risu Na, Siqin Tong, and Alu Si.
\newblock Himawari-8 satellite based dynamic monitoring of grassland fire in
  china-mongolia border regions.
\newblock {\em Sensors}, 18(1):276, 2018.

\bibitem{GiglioMODIS}
Louis Giglio, Wilfrid Schroeder, and Christopher Justice.
\newblock The collection 6 modis active fire detection algorithm and fire
  products.
\newblock {\em Remote Sensing of Environment}, 178:31--41, 06 2016.

\bibitem{chen2022california}
Yang Chen, Stijn Hantson, Niels Andela, Shane~R Coffield, Casey~A Graff,
  Douglas~C Morton, Lesley~E Ott, Efi Foufoula-Georgiou, Padhraic Smyth,
  Michael~L Goulden, et~al.
\newblock California wildfire spread derived using viirs satellite observations
  and an object-based tracking system.
\newblock {\em Scientific data}, 9(1):1--15, 2022.

\bibitem{flasse1996contextual}
SP~Flasse and P~Ceccato.
\newblock A contextual algorithm for avhrr fire detection.
\newblock {\em International Journal of Remote Sensing}, 17(2):419--424, 1996.

\bibitem{zhang2017approaches}
Tianran Zhang, Martin~J Wooster, and Weidong Xu.
\newblock Approaches for synergistically exploiting viirs i-and m-band data in
  regional active fire detection and frp assessment: A demonstration with
  respect to agricultural residue burning in eastern china.
\newblock {\em Remote Sensing of Environment}, 198:407--424, 2017.

\bibitem{xu2017major}
Weidong Xu, Martin~J Wooster, Takayuki Kaneko, Jiangping He, Tianran Zhang, and
  Daniel Fisher.
\newblock Major advances in geostationary fire radiative power (frp) retrieval
  over asia and australia stemming from use of himarawi-8 ahi.
\newblock {\em Remote Sensing of Environment}, 193:138--149, 2017.

\bibitem{di2018geostationary}
Valeria Di~Biase and Giovanni Laneve.
\newblock Geostationary sensor based forest fire detection and monitoring: An
  improved version of the sfide algorithm.
\newblock {\em Remote Sensing}, 10(5):741, 2018.

\bibitem{wooster2021satellite}
Martin~J Wooster, Gareth~J Roberts, Louis Giglio, David~P Roy, Patrick~H
  Freeborn, Luigi Boschetti, Chris Justice, Charles Ichoku, Wilfrid Schroeder,
  Diane Davies, et~al.
\newblock Satellite remote sensing of active fires: History and current status,
  applications and future requirements.
\newblock {\em Remote Sensing of Environment}, 267:112694, 2021.

\bibitem{roberts2014development}
G~Roberts and MJ~Wooster.
\newblock Development of a multi-temporal kalman filter approach to
  geostationary active fire detection \& fire radiative power (frp) estimation.
\newblock {\em Remote Sensing of Environment}, 152:392--412, 2014.

\bibitem{filizzola2016rst}
Carolina Filizzola, Rosita Corrado, Francesco Marchese, Giuseppe Mazzeo,
  Rossana Paciello, Nicola Pergola, and Valerio Tramutoli.
\newblock Rst-fires, an exportable algorithm for early-fire detection and
  monitoring: Description, implementation, and field validation in the case of
  the msg-seviri sensor.
\newblock {\em Remote Sensing of Environment}, 186:196--216, 2016.

\bibitem{rostami2022active}
Amirhossein Rostami, Reza Shah-Hosseini, Shabnam Asgari, Arastou Zarei,
  Mohammad Aghdami-Nia, and Saeid Homayouni.
\newblock Active fire detection from landsat-8 imagery using deep multiple
  kernel learning.
\newblock {\em Remote Sensing}, 14(4):992, 2022.

\bibitem{xu2017real}
Guang Xu and Xu~Zhong.
\newblock Real-time wildfire detection and tracking in australia using
  geostationary satellite: Himawari-8.
\newblock {\em Remote Sensing Letters}, 8(11):1052--1061, 2017.

\bibitem{udahemuka2020characterization}
Gustave Udahemuka, Barend~J van Wyk, and Yskandar Hamam.
\newblock Characterization of background temperature dynamics of a
  multitemporal satellite scene through data assimilation for wildfire
  detection.
\newblock {\em Remote Sensing}, 12(10):1661, 2020.

\bibitem{haiyangchallenge}
Haiyang Chen.
\newblock Challenges and corresponding solutions of generative adversarial
  networks (gans): A survey study.
\newblock {\em Journal of Physics: Conference Series}, 1827:012066, 03 2021.

\bibitem{yang2022diffusion}
Ruihan Yang, Prakhar Srivastava, and Stephan Mandt.
\newblock Diffusion probabilistic modeling for video generation.
\newblock {\em arXiv preprint arXiv:2203.09481}, 2022.

\bibitem{mei2022vidm}
Kangfu Mei and Vishal~M. Patel.
\newblock Vidm: Video implicit diffusion models, 2022.

\bibitem{yan2021videogpt}
Wilson Yan, Yunzhi Zhang, Pieter Abbeel, and Aravind Srinivas.
\newblock Videogpt: Video generation using vq-vae and transformers.
\newblock {\em arXiv preprint arXiv:2104.10157}, 2021.

\bibitem{croitoru2023diffusion}
Florinel-Alin Croitoru, Vlad Hondru, Radu~Tudor Ionescu, and Mubarak Shah.
\newblock Diffusion models in vision: A survey.
\newblock {\em IEEE Transactions on Pattern Analysis and Machine Intelligence},
  2023.

\bibitem{song2019generative}
Yang Song and Stefano Ermon.
\newblock Generative modeling by estimating gradients of the data distribution.
\newblock {\em Advances in neural information processing systems}, 32, 2019.

\bibitem{ho2022classifier}
Jonathan Ho and Tim Salimans.
\newblock Classifier-free diffusion guidance.
\newblock {\em arXiv preprint arXiv:2207.12598}, 2022.

\bibitem{vae}
Diederik~P Kingma and Max Welling.
\newblock Auto-encoding variational bayes.
\newblock {\em arXiv preprint arXiv:1312.6114}, 2013.

\bibitem{ronneberger2015u}
Olaf Ronneberger, Philipp Fischer, and Thomas Brox.
\newblock U-net: Convolutional networks for biomedical image segmentation.
\newblock In {\em Medical Image Computing and Computer-Assisted
  Intervention--MICCAI 2015: 18th International Conference, Munich, Germany,
  October 5-9, 2015, Proceedings, Part III 18}, pages 234--241. Springer, 2015.

\bibitem{he2016deep}
Kaiming He, Xiangyu Zhang, Shaoqing Ren, and Jian Sun.
\newblock Deep residual learning for image recognition.
\newblock In {\em Proceedings of the IEEE conference on computer vision and
  pattern recognition}, pages 770--778, 2016.

\bibitem{wu2018group}
Yuxin Wu and Kaiming He.
\newblock Group normalization.
\newblock In {\em Proceedings of the European conference on computer vision
  (ECCV)}, pages 3--19, 2018.

\bibitem{cadena2021spade}
Pablo Rodrigo~Gantier Cadena, Yeqiang Qian, Chunxiang Wang, and Ming Yang.
\newblock Spade-e2vid: Spatially-adaptive denormalization for event-based video
  reconstruction.
\newblock {\em IEEE Transactions on Image Processing}, 30:2488--2500, 2021.

\bibitem{yolov5}
Ultralytics.
\newblock Yolov5 classification models, apple m1, reproducibility, clearml and
  deci.ai integrations.
\newblock {\em DOI: 10.5281/zenodo.7002879}, 2022.
\newblock \url{https://github.com/ultralytics/yolov5}.

\bibitem{mathieu2015deep}
Michael Mathieu, Camille Couprie, and Yann LeCun.
\newblock Deep multi-scale video prediction beyond mean square error.
\newblock {\em arXiv preprint arXiv:1511.05440}, 2015.

\bibitem{medel2016anomaly}
Jefferson~Ryan Medel and Andreas Savakis.
\newblock Anomaly detection in video using predictive convolutional long
  short-term memory networks.
\newblock {\em arXiv preprint arXiv:1612.00390}, 2016.

\bibitem{mahadevan2010anomaly}
Vijay Mahadevan, Weixin Li, Viral Bhalodia, and Nuno Vasconcelos.
\newblock Anomaly detection in crowded scenes.
\newblock In {\em 2010 IEEE computer society conference on computer vision and
  pattern recognition}, pages 1975--1981. IEEE, 2010.

\bibitem{liu2018future}
Wen Liu, Weixin Luo, Dongze Lian, and Shenghua Gao.
\newblock Future frame prediction for anomaly detection--a new baseline.
\newblock In {\em Proceedings of the IEEE conference on computer vision and
  pattern recognition}, pages 6536--6545, 2018.

\bibitem{burgeLSTM}
John Burge, Matthew Bonanni, Matthias Ihme, and Lily Hu.
\newblock Convolutional lstm neural networks for modeling wildland fire
  dynamics.
\newblock 12 2020.

\end{thebibliography}

\end{document}